\documentclass[lettersize, journal]{IEEEtran}
\usepackage{multirow}
\usepackage{color}
\usepackage{hyperref}
\usepackage[normalem]{ulem}
\usepackage{balance}
\usepackage{cases}
\usepackage{array}
\usepackage{bbding}
\usepackage{amssymb}
\usepackage{graphicx}
\usepackage{booktabs}
\usepackage{diagbox}
\usepackage{caption}
\usepackage{graphicx}
\usepackage{subcaption}

\usepackage{orcidlink}

\usepackage{tabularx}
\usepackage{subcaption}
\usepackage{multirow}
\usepackage{multicol}
\usepackage{algorithm}
\usepackage{algorithmic}
\usepackage{booktabs} 
\usepackage{listings}
\DeclareCaptionStyle{ruled}{labelfont=normalfont,labelsep=colon,strut=off} 
\floatstyle{ruled}
\newfloat{listing}{tb}{lst}{}
\floatname{listing}{Listing}

\begin{document}

\title{AwesomeMeta+: A Mixed-Prototyping Meta-Learning System Supporting AI Application Design Anywhere}

\author{Jingyao~Wang,
        Yuxuan~Yang,
        Wenwen~Qiang,
        Changwen~Zheng,
        and Fuchun~Sun,~\IEEEmembership{Fellow,~IEEE}
\IEEEcompsocitemizethanks{\IEEEcompsocthanksitem J. Wang, W. Qiang, and C. Zheng are with University of Chinese Academy of Sciences, Beijing, China. They are also with National Key Laboratory of Space Integrated Information System, Institute of Software Chinese Academy of Sciences, Beijing, China. E-mail: {wangjingyao2023, qiangwenwen, changwen}@iscas.ac.cn.
\IEEEcompsocthanksitem  Y. Yang is with Nanjing Forestry University, Nanjing, China. E-mail: qingzhuo0409@gmail.com.
\IEEEcompsocthanksitem F. Sun is with Tsinghua University, Beijing, China. E-mail: 	fcsun@tsinghua.edu.cn.
}}

\markboth{IEEE Transactions on Automation Science and Engineering}%
{Wang \MakeLowercase{\textit{et al.}}: AwesomeMeta+: A Mixed-Prototyping Meta-Learning System Supporting AI Application Design Anywhere}


\maketitle

\begin{abstract}
Meta-learning, also known as ``learning to learn'', enables models to acquire great generalization abilities by learning from various tasks. Recent advancements have made these models applicable across various fields without data constraints, offering new opportunities for general artificial intelligence. However, applying these models can be challenging due to their often task-specific, standalone nature and the technical barriers involved. To address this challenge, we develop AwesomeMeta+, a prototyping and learning system designed to standardize the key components of meta-learning within the context of systems engineering. It standardizes different components of meta-learning and uses a building block metaphor to assist in model construction.  
By employing a modular, building-block approach, AwesomeMeta+ facilitates the construction of meta-learning models that can be adapted and optimized for specific application needs in real-world systems. The system is developed to support the full lifecycle of meta-learning system engineering, from design to deployment, by enabling users to assemble compatible algorithmic modules. We evaluate AwesomeMeta+ through feedback from 50 researchers and a series of machine-based tests and user studies. The results demonstrate that AwesomeMeta+ enhances users' understanding of meta-learning principles, accelerates system engineering processes, and provides valuable decision-making support for efficient deployment of meta-learning systems in complex application scenarios.
\end{abstract}


\section{Introduction}
\label{sec:1}

\IEEEPARstart{T}{he} goal of machine learning (ML) is to make models solve not just simple reactive tasks (associating stimuli with instant feedback), but also more complex cognitive tasks (requiring inference and response time), thus interacting with the external world \cite{jordan2015machine}. Recent methods have taken off with a profound impact on various fields, e.g., economy, society, industry, and education \cite{zhang2021study, liu2020adaptive, kaul2020history, chen2020artificial}. The researchers designed them to solve different tasks \cite{gao2023survey,bhatia2010survey}, e.g., Mask RCNN \cite{maskrcnn} is mostly used for image segmentation, and Diffusion \cite{diffusion} is mostly used for image generation. Although existing task-specific methods can surpass human intelligence in some tasks, they rely on large amounts of data or deep networks with sufficient training. 
The emergence of meta-learning and related models provides an opportunity to solve the above issues. It can endow machine learning with the ability to adapt to new developments like humans and complete multiple tasks that do not rely on human experience. Meta-learning \cite{finn2019online,hospedales2021meta} can realize the idea of autonomous driving on roads with unknown conditions, or allow a robotic arm to handle various objects of different specifications and weights, achieving outstanding performance in multiple scenarios \cite{peng2002improved,rivolli2022meta}.

However, despite the excellent performance of current meta-learning frameworks across various fields, integrating meta-learning models into large-scale systems or engineering projects remains a significant challenge. 
One primary issue is the inherent complexity of deploying meta-learning frameworks, which differ substantially from traditional machine learning approaches. Meta-learning involves multi-layer learning, where tasks serve as the fundamental units, and integrates diverse algorithms such as optimization-based, model-based, metric-based, and Bayesian methods \cite{wiering2012reinforcement, raghu2019rapid}. Additionally, tasks span various domains, including reinforcement learning, few-shot learning, and multi-modal learning \cite{levine2020offline, botvinick2019reinforcement, sun2021research}, necessitating the use of advanced optimization techniques like implicit gradient methods \cite{implicit1, implicit2, implicit3, implicit4, implicit5}, differentiable proxies \cite{differentiable1,differentiable2,differentiable3,differentiable4}, and single-layer approximations \cite{reptile,metaaug,nichol2018first,anil}. These factors significantly increase the complexity of model deployment and integration. Moreover, given that meta-learning is a relatively novel field compared to traditional areas like computer vision and natural language processing, there is a limited community and few available tools for model deployment, raising the application threshold and prolonging deployment time.
Furthermore, many open-source meta-learning frameworks are isolated and tailored for single tasks or standalone applications. For instance, in autonomous driving, engineers may use ReBAL \cite{nagabandi2018learning} for route planning or ProtoNet \cite{protonet} for scalable features in multilingual translation tasks. To unify the outcomes of these models within a single system, engineers must manually create interfaces between disparate frameworks, optimizing parallel efficiency across diverse model environments. Additionally, despite the availability of numerous models, designers often rely on a small subset or a single widely-used model, potentially missing out on leveraging features from lesser-known models that could enhance system performance.

\begin{figure*}
  \centering
    \begin{subfigure}{0.65\textwidth}
         \centering
         \includegraphics[width=\textwidth]{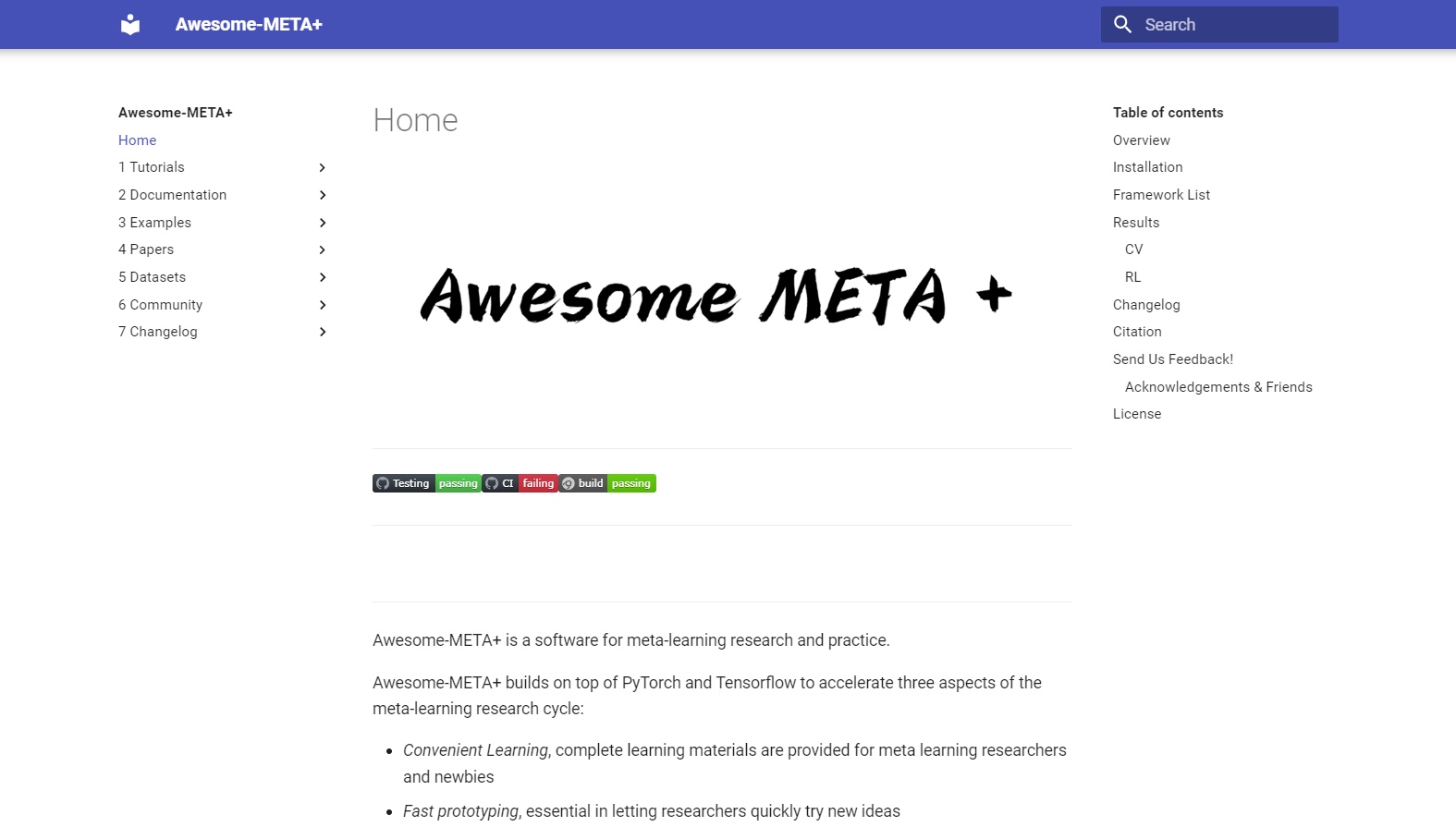}
         \caption{Web on laptop computer.}
     \end{subfigure}
    \hspace{0.5in}
     \begin{subfigure}{0.225\textwidth}
         \centering
         \includegraphics[width=\textwidth]{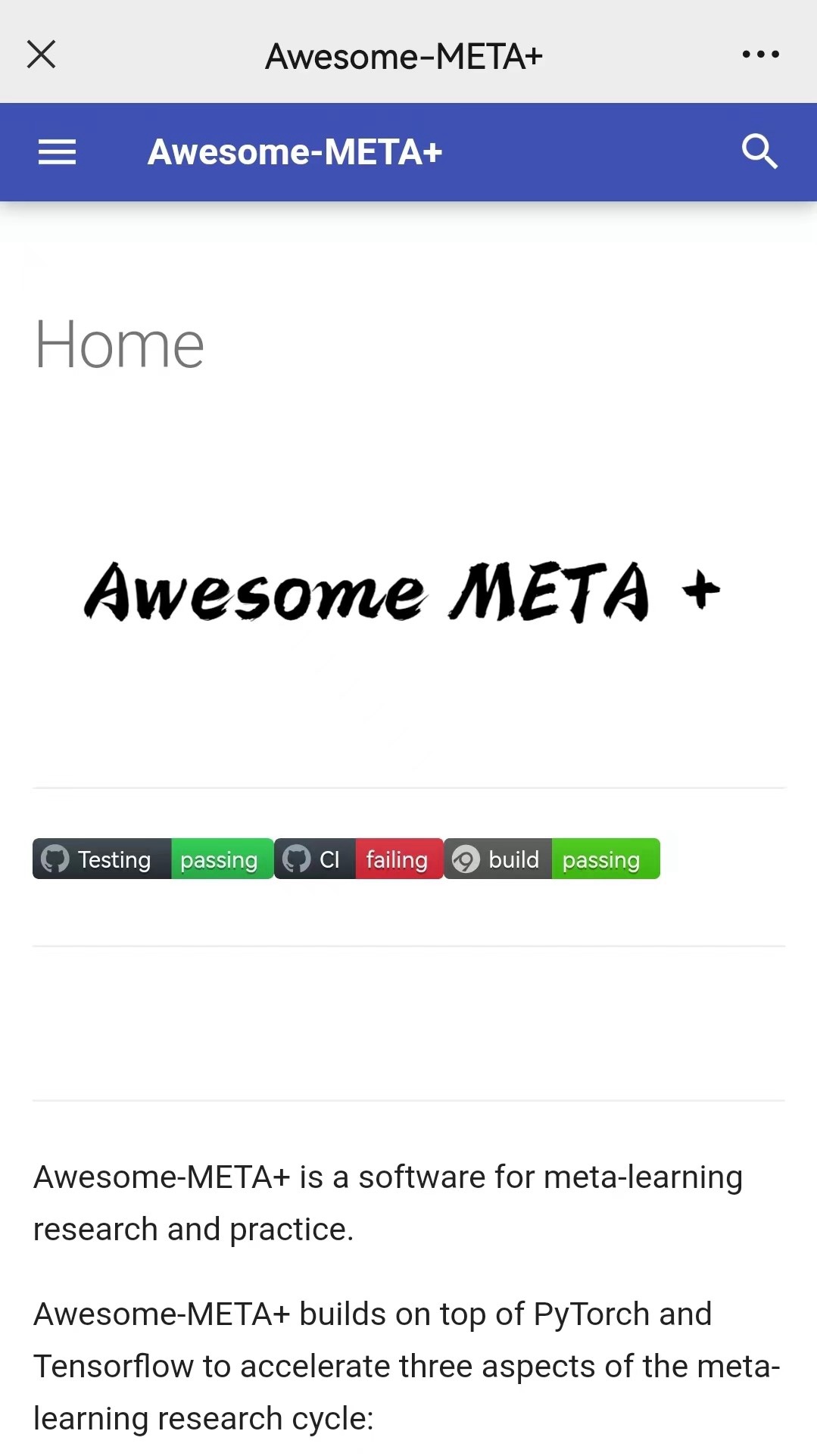}
         \caption{Web on mobile devices}
     \end{subfigure}
    \caption{Web of Awesome-META+, A Prototyping and Learning System. \textbf{Left:} Homepage for PC. \textbf{Right:} Homepage for mobile devices. AwesomeMeta+ uses a responsive interactive web page that can be opened on any device and perform global searches.}
    \label{home}
\end{figure*}

To gain a comprehensive understanding of the challenges associated with applying existing meta-learning models, we conducted a formative study involving 50 researchers, identifying four key obstacles in model selection, accessibility of learning resources, prototyping efficiency, and cross-domain model integration. Furthermore, during system development, we incorporated test ports to supplement our findings and iteratively refine our design objectives. First, users often lack awareness of the applicability and functional scope of different meta-learning models, making model selection and adaptation difficult. Second, acquiring relevant learning materials is a cumbersome process, as literature searches frequently yield irrelevant or outdated resources. Third, prototyping with meta-learning models is inefficient due to limited publicly available code and the deployment challenges posed by environment dependencies and version incompatibilities. Finally, even after deployment, integrating models across diverse fields and modalities remains problematic. Current implementations predominantly rely on serial execution or single-algorithm deployment, leading to inefficiencies. Additionally, complex hyperparameter tuning during model deployment further impacts performance.

In response to the above challenges, we develop AwesomeMETA+, a prototyping and learning system designed to lower the technical barriers of meta-learning through standardization, modularization, and integration of meta-learning frameworks. Its design adheres to four core objectives, each addressing one of the identified challenges. Firstly, for model selection and adaptability, AwesomeMETA+ integrates various categories of meta-learning algorithms and provides scenario-based recommendations and performance analysis. Users can select suitable models and deployment strategies based on structured system case modules. Considering learning resource accessibility, based on feedback from eight scholars, the system curates and organizes meta-learning resources for different expertise levels. A web-based learning platform with periodic updates ensures users have continuous access to relevant materials. Besides, AwesomeMETA+ standardizes meta-learning components, including task construction, forward propagation, and gradient optimization, unifying disparate frameworks into a cohesive system. Users can construct models through a modular ``building block'' approach or select pre-assembled model packages for one-click deployment. The system also introduces a parameter search module to optimize model configurations, improving deployment efficiency across different fields and modalities.

To assess the effectiveness of Awesome-META+, we conduct machine-based performance testing and user studies. The system is evaluated for response speed and robustness across various use cases, while user studies measured engineers’ ability to deploy algorithm workflows in both structured application scenarios and exploratory settings. Results demonstrate that Awesome-META+ provides an intuitive mechanism for efficiently integrating and deploying meta-learning models across diverse domains, significantly enhancing usability and performance.
The main contributions of this work include:
\begin{itemize}
    \item We conduct a formative study with 50 researchers to identify the challenges when deploying meta-learning models. At the same time, we built a learning platform for different stages based on the advice of professionals, including literature, videos, and case extensions to help understand meta-learning technology.
    \item We conduct AwesomeMeta+, a prototype and learning system that standardizes, modular, and integrates different components of meta-learning to help engineers build models that adapt to any task according to their needs. AwesomeMeta+ also provides convenient deployment solutions, various application cases, and application analysis for different methods to advance the engineering process.
    \item We conduct machine-based testing and user studies to evaluate the effectiveness of AwesomeMeta+, which demonstrates that AwesomeMeta+ achieves stable service and accelerates the engineering process.
\end{itemize}


\section{Related Work}
In this section, we provide a concise overview of key developments in meta-learning (Subsection \ref{sec:2.1}) and human-AI interaction (Subsection \ref{sec:2.2}), both of which inform the proposed mixed-prototyping meta-learning system taken in this research.

\subsection{Meta-Learning}
\label{sec:2.1}
Meta-learning facilitates rapid adaptation to novel tasks using only a few training samples, thus reducing the need for comprehensive retraining and offering solutions in data-scarce or resource-constrained contexts \cite{vanschoren2018meta, hospedales2021meta}. It is commonly categorized into four methodological families:
\textbf{(i) Optimization-based methods} learn an optimal initialization that enables quick convergence for new tasks, such as fault diagnosis \cite{feng2022semi} or mitigating catastrophic forgetting \cite{chi2022metafscil}. Classic examples include MAML \cite{finn2017model}, Reptile \cite{nichol2018reptile}, and MetaOptNet \cite{lee2019meta}.
\textbf{(ii) Model-based methods} emphasize architectural adaptations or specialized memory structures to facilitate rapid updates, exemplified by MANNs \cite{santoro2016meta}, NTMs \cite{graves2014neural}, and SNAIL \cite{mishra2017simple}.
\textbf{(iii) Metric-based methods} learn embedding functions that map instances from different tasks into a space suitable for straightforward classification via non-parametric techniques. Representative works include Siamese Network \cite{koch2015siamese}, Matching Network \cite{vinyals2016matching}, Prototypical Network \cite{snell2017prototypical, cheng2022holistic}, Relation Network \cite{sung2018learning}, and GNN-based approaches \cite{hospedales2021meta}.
\textbf{(iv) Bayesian-based methods} leverage conditional probability to guide the meta-learning process, as illustrated by CNAPs \cite{zhang2021shallow}, SCNAP \cite{bateni2020improved}, BOOM \cite{grant2018recasting}, and VMGP \cite{myers2021bayesian}.
Despite their varied application contexts, these approaches are amenable to a common set of modular components (e.g., task distribution, meta-learner, base learner, meta-loss function, update strategy). Such abstraction plays a pivotal role in systems engineering, as it enables standardization across distinct meta-learning frameworks and facilitates integration into broader pipelines. Building on this, the present work systematically unifies these components so that users can configure their own frameworks in a ``building block'' fashion. For instance, the meta-learning optimization component includes strategies such as implicit gradient, differentiable proxies, and single-layer approximations, each accompanied by supporting performance analyses and real-world use cases to guide selection.

\subsection{Human-AI Interaction for System Engineering}
\label{sec:2.2}

With the rapid development of artificial intelligence (AI) technology, human-AI interaction systems have been applied in various fields, such as autonomous driving \cite{wang2022social,schneemann2016analyzing} and image design \cite{nie2023application,wang2023cssl}. They not only focuses on how to improve the performance of AI systems, but also how to design, optimize, and understand the interaction between humans and AI. Many studies \cite{bisen2022responsive,wang2022research,duan2021gesture} in recent years have explored the way humans perceive and interact with AI. 
Lin et al. \cite{lin2024jigsaw} constructed a formative study and found that AI can provide humans with new perspectives and design exploration paths. 
Lv et al. \cite{lv2022deep} conducted a case study on human-computer interaction systems and found that human-computer interaction systems accelerated the engineering and application process and provided engineers with a more convenient exploration path. This study follows this idea and builds a tool to help humans understand and use AI, while also prototyping workflows for new applications or scenarios with the support of AI.

Note that the construction of human-AI interaction systems are still challenging \cite{related21,related22,related23,related24}, which is widely found in the community. Subramonyam et al. \cite{subramonyam2021towards} argue that the challenge of using AI as design material is that the properties of AI only appear in a part of the user experience design. Yang et al. \cite{yang2020re} found that designers often find it difficult to design with AI because they are uncertain about the capabilities of AI and the complexity of AI outputs. In this work, we provide mechanisms to help designers overcome these challenges, such as providing model performance results in case scenarios and supporting easy inspection and manipulation of AI outputs using real data.
In addition, Gmeiner et al. \cite{gmeiner2023exploring} found that the main challenges faced by users when using AI tools for collaborative creation are understanding and operating AI outputs and the difficulty in adapting and compatibility between algorithms. Choe et al. \cite{choe2022betty} conducted case studies on different classic models and found that the frameworks of different models are very different, and version isolation makes user deployment difficult. Therefore, we standardize and modularize different categories of meta-learning algorithms and integrate them all into the provided online environment, so that users can deploy their own models based on compatible modules.
Furthermore, in order to facilitate users to understand meta-learning-related technologies, we provide a large amount of learning materials and literature through formative studies based on the feedback and suggestions of multiple researchers. At the same time, we adopt responsive interactive design, so that users can search for the required information on any device, e.g., laptop computer and mobile devices. As far as we know, this study is the first that has achieved the standardization, modularization, and integration of the specific field of meta-learning, allowing users to learn, understand, and apply models in various scenarios.


\section{Formative Study}
\label{sec:3}
We conduct a formative study for 50 volunteers, including researchers from different fields, to understand how users try to use AI models in their work while collecting information for building new tools to support creative works based on meta-learning. In this section, we first introduce the procedure and participants of the formative study. Next, we perform requirements analysis and system design based on the feedback and suggestions we received.

\subsection{Procedure and Participants}
In the formative study, we recruit 50 volunteers, including 23 women and 27 men, aged between 23 and 52. They come from different professional fields, 36 of whom were from the computer field, and the remaining 14 from various fields including interior design, astronomy, medicine, electronics, etc. All participants have more than three years of engineering and scientific research experience and used AI tools or simple neural networks to support the implementation of their research and projects, and 12 of them had more than 10 years of work experience. We first collect the goals that each participant hope to achieve with the meta-learning models and the difficulties they currently encounter based on their one-minute description. Furthermore, we conduct a survey on the needs of "learning" and "application" to clarify the services that participants hope AwesomeMETA+ can provide. After the survey, we collect feedback from the participants, conduct a demand analysis, and summarize the challenges and barriers faced by the current application of meta-learning. Note that after building AwesomeMETA+, we provide participants with a usage portal to evaluate whether it met users' needs.

\subsection{Requirements Analysis and System Design}
Based on the feedback from the above formative study, we conduct requirements analysis and sort out the functional components of AwesomeMETA+ to meet these needs. Specifically, we first analyze the user profile (Subsection \ref{sec:3.1}) and applicable scenarios (Subsection \ref{sec:3.2}), and then design the intended function (Subsection \ref{sec:3.3}) and acceptance criteria (Subsection \ref{sec:3.4}) to help us build AwesomeMeta+.

\subsubsection{User Profile}
\label{sec:3.1}
AwesomeMeta+ is designed to facilitate learning and applying meta-learning models. The target users include individuals who are interested in or working in the field of meta-learning, as shown in Table \ref{tab:Audience groups}. The system is specifically divided into the following three groups based on the potential needs of users.

\begin{table*}[h]
\centering
\caption{User Profile.}
\label{tab:Audience groups}
\begin{tabularx}{\textwidth}{l|X}
\toprule
\textbf{No.} & \textbf{Description} \\
\midrule
Group 1 & Scholars or practitioners in the field of meta-learning.\\
\midrule
Group 2 & Beginners interested in the field of meta-learning. \\
\midrule
\multirow{2}{*}{Group3} & Scholars and industry practitioners in various fields hope to use the meta-learning paradigm to improve framework performance or apply it to landing products. \\
\bottomrule
\end{tabularx}
\end{table*}

\subsubsection{Application Scenarios}
\label{sec:3.2}
Based on the feedback from the formative study above, AwesomeMeta+ needs to include four scenarios to target different user groups and needs. Table \ref{tab:Application Scenarios} describes the users and their needs for each scenario.

\begin{table*}[h]
\centering
\caption{Application Scenarios}
\label{tab:Application Scenarios}
\begin{tabularx}{\textwidth}{l|l|X}
\toprule
\textbf{Scenario } & \textbf{Target Audience} & \textbf{Requirements} \\
\midrule
\multirow{2}{*}{Scenario 1} & \multirow{2}{*}{Group 1/2/3}  & Users need to configure specific frameworks on their local machines and understand the core technologies and ideas behind the framework's code.\\
\midrule
\multirow{2}{*}{Scenario 2} &\multirow{2}{*}{Group 1} & Academic researchers need to conduct comparative experiments on multiple meta-learning frameworks to obtain data or improve the performance of specific tasks.\\
\midrule
\multirow{2}{*}{Scenario 3} &\multirow{2}{*}{Group 2} & Individuals who want to understand the current development status of meta-learning, engage in systematic learning, and obtain relevant materials. \\
\midrule
\multirow{2}{*}{Scenario 4} & \multirow{2}{*}{Group 3} & Users hope to use meta-learning to complete multiple specific tasks in fields such as reinforcement learning and achieve industrial applications.\\
\bottomrule
\end{tabularx}
\end{table*}

\subsubsection{Intended Function}
\label{sec:3.3}

We should build a system that makes users able to search for needed literature, deploy models, access learning resources, and transfer tasks across different domains. The system is designed to be user-friendly and easy to navigate. It provides a range of resources and tutorials to help users learn more about meta-learning and related topics. Meanwhile, due to the isolation between different models, it is necessary to unify various types of meta-learning methods. To achieve this, we abstract meta-learning as a method composed of multiple fundamental modules, such as task construction, meta-learner, base learner, optimizer, and others, while providing selectable implementations for each module. Ultimately, by standardizing and modularizing meta-learning methods and integrating them into the system, users will be able to quickly build meta-learning models for different tasks, akin to assembling building blocks. Table \ref{tab:functional-points} provide the brief illustration of intended functions that the conducted system need have.

\begin{table*}
\centering
\caption{Intended Functions.}
\label{tab:functional-points}
\begin{tabularx}{\textwidth}{l|l|X}
\toprule
\textbf{Functional point ID} & \textbf{Function Name} & \textbf{Description} \\
\midrule
\multirow{5}{*}{Function 1} & \multirow{5}{*}{Search Functionality} & Users can locate the desired meta-learning framework, paper, and other related information by using the search bar or navigating through the menu bar, including "Home", "Tutorials", "Documentation", "Examples", "Papers", "Datasets", "Community", "Changelog", and "GitHub". Each module contains multiple sub-search options (e.g., "Changelog" shows version iteration information). \\
\midrule
\multirow{6}{*}{Function 2} & \multirow{6}{*}{Model Deployment} & Users can browse the frameworks, models, and datasets provided by the platform on the "Home" page and then locate them according to their needs through four ways: (i) directly entering keywords in the search bar; (ii) clicking to the corresponding framework deployment method on the "Home" page and obtaining specific details from modules such as "Tutorials"; (iii) pull the source code of meta-learning frameworks and deployment details with one click for local testing; and (iv) build their own model by calling standardized modules that integrated into the system. \\
\midrule
\multirow{3}{*}{Function 3}  & \multirow{3}{*}{Learning Platform}  & Users can locate the "Papers", "Datasets", and "Community" modules according to their needs, and obtain resources such as the learning curve of the platform, as well as links to download relevant blogs, monographs, and papers. \\
\midrule
\multirow{6}{*}{Function 4}  & \multirow{6}{*}{Multi-Domain Task Transfer} & Users can use the "Tutorials", "Documentation", and "Examples" modules to learn about the system's usage instructions and framework information, different domain tasks corresponding to framework details and optimization ideas, and actual cases (such as performance comparison of various frameworks in small-sample image classification using metrics such as ACC, AP, etc.) to locate their desired goals and complete configuration. \\
\midrule
\multirow{3}{*}{Function 5}  & \multirow{3}{*}{Feedback}  & Users can write feedback or suggestions in the feedback section on the system's main page "Home" (which actually redirects to GitHub's issues) for future maintenance or adding new learning materials.\\
\midrule
\end{tabularx}
\end{table*}

\subsubsection{Acceptance Criteria}
\label{sec:3.4}
To better meet the needs of users, we need to achieve four goals, i.e., reliability, ease of use, maintainability, and iterative updates. Table \ref{tab:Acceptance criteria} provides brief descriptions of these goals.

\begin{table*}[h]
\centering
\caption{Acceptance Criteria.}
\label{tab:Acceptance criteria}
\begin{tabularx}{\textwidth}{l|X}
\toprule
\textbf{Goal} & \textbf{Specifics} \\
\midrule
\multirow{2}{*}{Reliability}  & The system's framework has been extensively tested and validated through experiments with ample data. All modules have been tested to ensure their reliability.\\
\midrule
\multirow{2}{*}{Ease of Use} & The user interface is simplified to enable users to quickly and accurately find the required information and deploy the framework quickly. \\
\midrule
Maintainability & User feedback are promptly addressed, and the system is maintained within one week. \\
\midrule
\multirow{2}{*}{Updates and Iterations} & The system is designed to allow developers to quickly iterate and add necessary modules based on the current software ecosystem. A "Changelog" is specifically set up within the system to view product updates and iteration information.\\
\bottomrule
\end{tabularx}
\end{table*}


\section{System Development}
\label{sec:4}
Awesome-META+ is a meta-learning prototype and learning system that will be made available to a wide range of internet users. 
It has two forms: web page and installation package. The former is mainly used for online learning, application cases, etc., while the latter is mainly used to build users' own models based on standardized meta-learning modules of system integration.
The system comprises nine modules, including "Home", "Tutorials", "Documentation", "Examples", "Papers", "Datasets", "Community", "Changelog", and "GitHub", covering everything that is required for the application of typical meta-learning frameworks. Figure \ref{home} shows the homepage of Awesome-Meta+.

These modules include deployment, usage tutorials, source code, practical examples, as well as academic materials related to meta-learning, such as papers, datasets, blogs, and video tutorials. Additionally, the system features multiple modules for updating logs and community building.

The system's website and code are hosted on GitHub+Vercel, and although the relevant repositories are currently private, they will be made public once the domain formalities are completed.

Users can click on different modules in the menu bar based on their needs, navigate to the corresponding webpages, and complete the relevant operations. Users can also search for models, methods, and practical examples of their interest to enrich the meta-learning research community.

The development and functions of this system consist of:
\begin{itemize}
    \item \textbf{Front-end System}(Section \ref{Front-end}): built with Python + Django + Material for MkDocs to create a responsive website that is bound to the meta-learning model, deployment, and academic information retrieval functions. The main interface, "Home," provides content and usage instructions for each page, with the design aimed at giving users a clear understanding of meta-learning and making it easy to use the Awesome-META+ system.
    
    \item \textbf{Algorithms and Deployment} (Section \ref{Algo}): code is written and reproduced using Python + Anaconda to build twelve typical frameworks, including those based on machine learning frameworks such as Pytorch/TensorFlow \cite{paszke2019pytorch, girija2016tensorflow}, seven benchmarks including CIFAR, and two training methods, distributed and single card. A modular design is used for easy developer rewriting, and online running examples are provided.
    
    \item \textbf{Information Integration} (Section \ref{Infor}): includes information crawling and push based on Python. High citation and click-through papers, "meta-learning" keyword papers from large conferences such as ICLR and ICML, and blogs are crawled, scored, and sorted, and then uploaded to the system.
    
    \item \textbf{Testing and Deployment} (Section \ref{Verification}): includes both automated and manual testing. Automated testing includes website function testing, access and API response testing, and automated testing of frameworks and algorithms. Deployment is done via GitHub + Vercel for web demo deployment.
\end{itemize}

\subsection{Front-end system}
\label{Front-end}
\begin{figure*}
    \centering
    \includegraphics[width=0.9\textwidth]{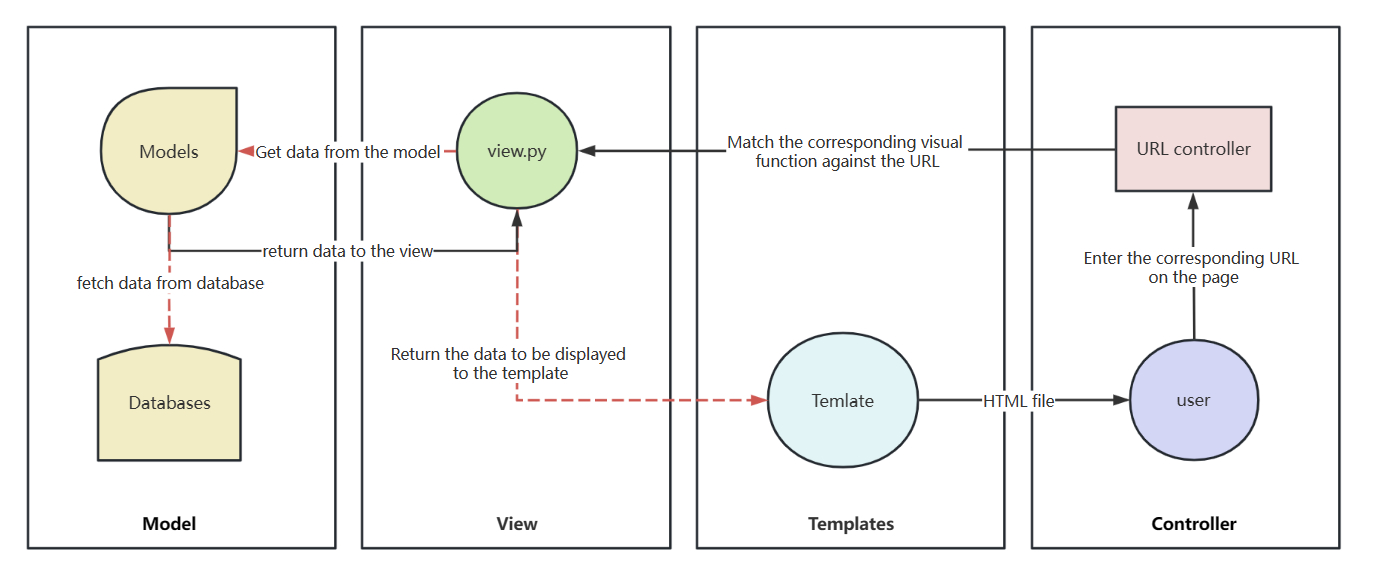}
    \caption{The process of web page interaction.}
    \label{fig:front-end}
\end{figure*}

The front-end of Awesome-META+ mainly consists of a user interaction interface, which is developed using Python and Django frameworks \cite{forcier2008python}, as well as Material for MkDocs. We introduce responsive web design and multiple ports required for model deployment to support local deployment and learning of the meta-learning framework. In addition, we provide practical examples for users to perform online training on platforms such as Colab \cite{gunawan2020development}. Figure \ref{fig:front-end} shows the process of web page interaction.

Django is an open-source web application framework written in Python. With Django, developers can implement a complete website with simple code and further develop a full-featured web service. Django is based on the MVC (Model-View-Controller) design pattern, consisting of three components: Model, View, and Controller. In addition, a URL dispatcher is required to distribute page requests to different Views, which then call the corresponding Model and Template. Python+Django web development has advantages such as low coupling, fast development, high reusability, and low maintenance costs, making it possible to quickly deploy a website.

To maintain code compatibility and maintainability, while considering the characteristics of the Awesome-META+ system and the time required for the deployment of the meta-learning framework, we choose to use Python for responsive interface development. This enables users to use the system on any mobile or PC interface, improving the system's applicability and ease of use.

\subsection{Algorithms and deployment}
\label{Algo}
In order to meet the needs of users as much as possible and achieve the purpose of meta-learning research, we conduct research from the perspectives of framework expansion, standardization, universality, and rapid deployment

\paragraph{Framework Expansion}
Before model development, we need to understand the meta-learning process from the perspective of software development and coding. Meta-learning is the process of learning how to learn, which refers to training machine learning algorithms that can automatically adapt to new tasks and environments. The general meta-learning process is as follows:
\begin{itemize}
  \item [1)] 
  Select meta-learning algorithm and model: Choose the appropriate meta-learning algorithm and model, such as gradient-based meta-learning, Bayesian meta-learning, meta-reinforcement learning, etc.      
  \item [2)]
  Select task distribution: Select several tasks from the given task distribution. For example, in image classification tasks, we conduct multiple 5-way 1-shot classification tasks with an adaptive task sampler \cite{wang2024towards}, where 5-way 1-shot means that the task contains five samples randomly collected from five classes, where 1 sample per class.
  \item [3)]
    Divide the select data: For all the selected tasks, divide the tasks into a training set and a test set. For each task, divide its data into a support set and a query set.
  \item [4)]
    Train the model: For each task in the training set, train the meta-learning model on the support set once to obtain the task-specific model for that task.
  \item [5)]
    Evaluate the model: Use the query set of that task to evaluate the trained task-specific model and obtain the performance metrics for that task.
  \item [6)]
    Update the meta-learning model: Use the performance metrics of the tasks in the same batch as the input to update the parameters of the meta-learning model, so that it can better adapt to the learning process of different tasks.
  \item [7)]
    Obtain the optimal model: Repeat Steps 4-6 above and train until the meta-learning model converges.
  \item [8)]
    Test the model: Use the test set to test the obtained meta-learning model for each task and obtain the final performance. If the model performs well, then it can be used to solve downstream tasks.
  \item [9)]
    Apply the meta-learning model: For new tasks, use the trained meta-learning model to select the most suitable model and parameters based on the nature and characteristics of the task, and perform learning and prediction accordingly.
\end{itemize}

\begin{figure}
    \centering
    \includegraphics[width=\linewidth]{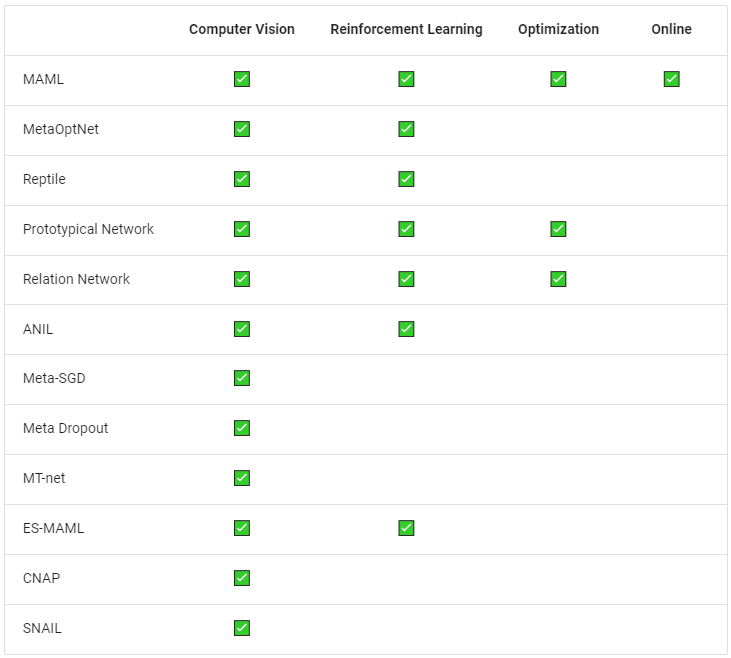}
    \caption{The framework resources provided by the Awesome-META+ system (Version 1.0).}
    \label{fig:framework}
\end{figure}

The above is the general meta-learning process, and we set up a parameter search module to support the fine-tuning of specific implementations.
We replicate and extend 12 meta-learning frameworks, where Figure \ref{fig:framework} lists the framework resources provided by the Awesome-META+ system. 
Taking MAML as an example, this model is a classic work in meta-learning, which uses optimization-based model training rules. 
During training, each batch consists of randomly sampled $N$-way $K$-shot tasks, where $N$ represents the number of classes per task, and $K$ represents the number of samples per class. The samples within each task are split into a support set and a query set. MAML is then trained using a bi-level optimization approach ~\cite{wang2021bridging,maml}. Specifically, in the inner loop, the model for each task is fine-tuned on the support set using the meta-learning model. In the outer loop, the meta-learning model is updated by learning from the query sets of all training tasks and their corresponding expected models. Consequently, it is widely hypothesized that as training progresses, the meta-learning model will gain more transferable knowledge, leading to improved performance on downstream tasks ~\cite{rivolli2022meta}.
This framework is compatible with any model trained by gradient descent and is suitable for various learning problems. In the development of our system, we use MAML as a typical example for multi-dataset scenarios and provide standardized instructions. Additionally, we configure multiple datasets, provide both PyTorch and TensorFlow code formats, support distributed and single-card training, and apply it to multiple scenarios such as reinforcement learning \cite{kaelbling1996reinforcement, li2017deep, dayan2008reinforcement}, gesture recognition \cite{wu1999vision, oudah2020hand}, and animal detection \cite{tian2020rethinking}. All this information and the corresponding learning materials can be found on the responsive web page of AwesomeMeta+.

\paragraph{Standardization}
We standardize the reproduced meta-learning framework, including task settings and optimization strategy (e.g., SGD-based second-order derivatives, implicit gradient, differentiable proxies, etc.), achieving deployment with unified settings. We also add these explanations and specific standardization content to the "Documentation" module on the Awesome-META+ system. The list of standardization modules and options is provided in the Appendix.
\begin{itemize}
    \item [1)] 
    Dataset standardization: Given any input dataset, the training and test tasks can be easily generated according to users' needs. A set of tasks is created from the given dataset, which accepts a list of task transforms that define the types of tasks to be sampled from the dataset. These tasks are lazily sampled when indexed (or called using the \texttt{.sample()} method), and their descriptions are cached for later use. If the number of tasks is set to -1, TaskDataset will not cache task descriptions and will continuously sample new descriptions. In this case, the length of TaskDataset is set to 1. Meanwhile, we supplement the task sampler extension package based on \cite{wang2024towards} in the second version, allowing users to choose a uniform sampler, low task diversity score sampler, high task diversity score sampler, or adaptive sampler according to their needs.
    \item [2)] 
    Model standardization (using MAML as an example): Taking the optimization-based model of the MAML class as an example, this class wraps any \texttt{nn.Module} and expands it using the \texttt{clone()} and \texttt{adapt()} methods. For the first-order version of MAML (i.e., FOMAML), the \texttt{first-order flag} is set to True during initialization. In addition to reproducing the models based on standardization rules, we have also reproduced the performance testing experiments in the paper to ensure the correctness of the reproduction of the twelve frameworks.
\end{itemize}

\begin{listing}
\caption{Code for Dataset Standardization}%
\label{lst:dataset}%
\begin{lstlisting}
dataset (Dataset) - Dataset of data to obtain tasks.
task_transforms (list, optional, default=None) - List of task transformations.
num_tasks (int, optional, default=-1) - Number of tasks to generate.

dataset = awesomemeta.data.MetaDataset(MyDataset())
transforms = [
    awesomemeta.data.transforms.NWays(dataset, n=5),
    awesomemeta.data.transforms.KShots(dataset, k=1),
    awesomemeta.data.transforms.LoadData(dataset), 
    awesomemeta.data.transforms.DataSplit(t=7,v=3),]
taskset = TaskDataset(dataset, transforms, num_tasks=20000)
for task in taskset:
    X, y = task
\end{lstlisting}
\end{listing}

\begin{listing}%
\caption{Code for Model Standardization}%
\label{lst:model}%
\begin{lstlisting}
model (Module) - Module to be wrapped
lr_alpha (float) - Fast adaptation learning rate
lr_beta (float) - Meta learner learning rate
first_order (bool, optional, default=False) - Whether to use the first-order approximation of MAML. (FOMAML)
allow_unused (bool, optional, default=None) - Whether to allow differentiation of unused parameters. Defaults to allow_nograd
allow_nograd (bool, optional, default=False) - Whether to allow adaptation with parameters that have requires_grad = False

linear = awesomemeta.algorithms.optimizationbased.MAML(nn.Linear(20, 10), lr_alpha=0.01, lr_beta=0.01)
clone = linear.clone()
error = loss(clone(X), y)
clone.adapt(error)
error = loss(clone(X), y)
error.backward()
\end{lstlisting}
\end{listing}

\paragraph{Universality}
To better meet the needs of users in deploying models, we have rewritten the framework to match multiple datasets, training methods, and multi-tasking capabilities. Additionally, we have made modifications to the training methods and environment versions with consideration for the hardware configurations of future servers and local hardware resources available to users.
\begin{itemize}
    \item [1)] 
    Regarding training methods, the framework includes distributed options (supporting multi-GPU training for hardware configurations such as servers and host machines with multiple graphics cards) as well as single-GPU training (supporting GPU-based hardware systems).
    \item [2)] 
    Regarding environment versions, some frameworks (such as MAML and Prototypical Network) are offered both PyTorch and TensorFlow versions, supporting multiple training formats.
\end{itemize}

\begin{listing}%
\caption{Code for Multi-GPU Training}%
\label{lst:multiGPU}%
\begin{lstlisting}
args.gpu = gpu
    torch.cuda.set_device(gpu)
    args.rank = args.node_rank * ngpus + gpu
    device = torch.device('cuda:%d' % args.gpu)

    if args.dist == 'ddp':
        dist.init_process_group(
            backend='nccl',
            init_method='tcp://%s' % args.dist_address,
            world_size=args.world_size,
            rank=args.rank,
        )

        n_gpus_total = dist.get_world_size()
        assert args.batch_size % n_gpus_total == 0
        args.batch_size //= n_gpus_total
        if args.rank == 0:
            print(f'===> {n_gpus_total} GPUs total; batch_size={args.batch_size} per GPU')

        print(f'===> Proc {dist.get_rank()}/{dist.get_world_size()}@{socket.gethostname()}', flush=True)

\end{lstlisting}
\end{listing}

\paragraph{Rapid Deployment}
In order to enable users to quickly deploy the required meta-learning models locally (or online) to meet their application needs, we take the following measures:
\begin{itemize}
    \item [1)] 
    Code encapsulation for on-demand running with only two lines of code: Configuration parameters such as dataset and training method are directly written into the framework, while functional modules are encapsulated by the class. Users can directly select the "Tutorials" module of the Awesome-META+ system, or follow the deployment instructions downloaded locally for training and running. The whole process only involves two statements: environment configuration and the command for using the meta-learning framework.
    \item [2)]
    Multi-scenario transfer: Experiment examples for multiple scenarios are provided on the "Examples" module of the Awesome-META+ system, demonstrating the effectiveness and practicality of the main framework.
    \item [3)]
    Online demo: To facilitate online training, we have attempted to set up a port linked to Colab on the web page. A MAML notebook is provided in Colab, which is a free online system that supports free GPU and can perform a series of operations such as model training.
\end{itemize}

\subsection{Information integration}
\label{Infor}

Another core feature of the Awesome-META+ system is to collect academic information, allowing users to access cutting-edge work in the field of meta-learning, including resources such as journals, conferences, and major reports. The information integration process is shown in the Figure \ref{fig:Infor}. This includes:

\begin{figure*}
    \centering
    \includegraphics[width=0.85\textwidth]{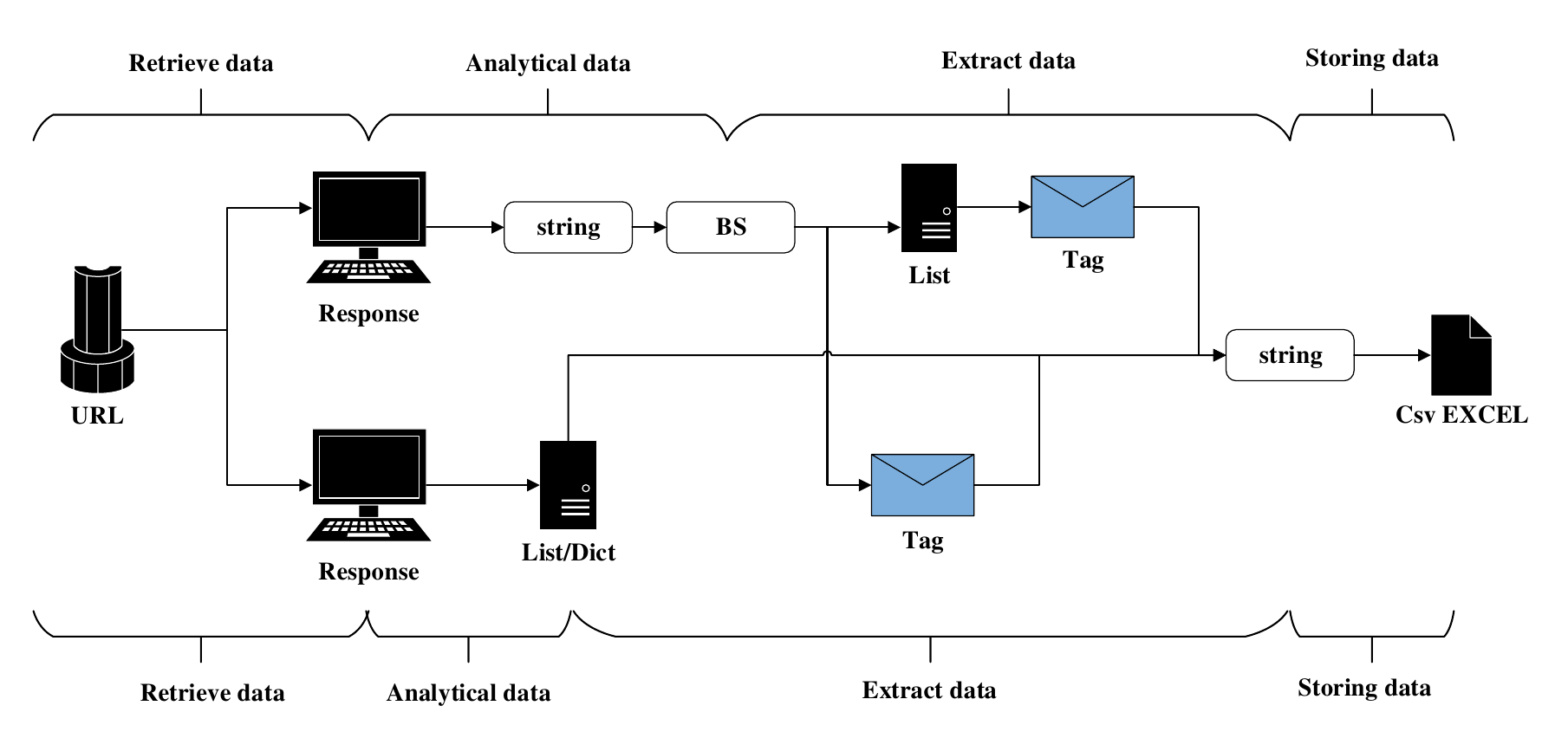}
    \caption{The process of information integration.}
    \label{fig:Infor}
\end{figure*}

\begin{itemize}
    \item [1)] 
    High-quality papers published in recent years: This metric is evaluated based on the impact factor of the journal, the rating of the conference, and the citation count of the paper itself. For example, the meta-learning works in international conferences such as NIPS, ICLR, and ICML are collected and pushed to the learning platform.
    \item [2)] 
    Works in the field of meta-learning: including introductory books for beginners and advanced books for practitioners, works are ranked based on their influence and recommendations from multiple blogs.
    \item [3)] 
    Discussion videos and conference links related to meta-learning: workshops at important conferences, etc.
    \item [4)] 
    Meta-learning-related blogs and videos: Blog resources include domestic websites, e.g., Stackflow. Video resources are mainly from websites, e.g., YouTube.
\end{itemize}

To achieve comprehensive information collection and summary, we perform the following work and then push the collected data into the system:
\begin{itemize}
    \item [1)]
    Information is crawled based on keywords for resource web pages and addresses, looking for information related to "meta-learning" and "learn-to-learn"; for works and reports, manual searches are conducted (based on discussions in forums).
    \item [2)] 
    The crawled information is sorted and filtered, including selecting the top 10 papers for each website based on paper citation counts (specifically, the citation count at conferences such as ICLR, oral presentations, and the number of collections on websites such as ResearchGate, based on the influence evaluation indicators provided by these websites) and finally retaining 30 papers; selecting 20 blogs and videos based on the number of likes and views, respectively.
    \item [3)] 
    The filtered information is summarized and uploaded to the system.
\end{itemize}

Finally, users can access paper information and related materials on the "Home" and "Papers" pages, and download or jump to the content they want to learn.

\begin{table*}
\begin{center}
\caption{Machine-based testing and evaluation of Awesome-META+. It shows the results of both automated testing and manual testing, with automated testing further divided into three aspects: (i) front-end web interaction, (ii) framework integration testing, and deployment, and (iii) information acquisition.}
\label{table:6}
\resizebox{0.9\linewidth}{!}{
\begin{tabular}{l|c|c|c}
\toprule
\textbf{Test (\%)} & \textbf{Front-end web interaction} & 
\textbf{Framework integration testing and deployment} & \textbf{Information acquisition} \\
\midrule
\textbf{Test 1} & 99.237 & Unit Tests (99.372) Functional Tests (97.832) & 100.000 \\
\textbf{Test 2} & 99.372 & Unit Tests (100.00) Functional Tests (99.382) & 99.827 \\
\textbf{Test 3} & 99.178 & Unit Tests (98.893) Functional Tests (99.478) & 98.728 \\
\textbf{Test 4} & 99.732 & Unit Tests (100.00) Functional Tests (100.00) & 99.387 \\
\textbf{Test 5} & 100.000 & Unit Tests (100.00) Functional Tests (99.974) & 100.000 \\
\textbf{Test 6} & 98.947 & Unit Tests (99.287) Functional Tests (98.237) & 98.728 \\
\textbf{Test 7} & 99.238 & Unit Tests (100.00) Functional Tests (97.473) & 100.000 \\
\textbf{Test 8} & 100.000 & Unit Tests (100.00) Functional Tests (100.00) & 100.000 \\
\textbf{Test 9} & 100.000 & Unit Tests (99.783) Functional Tests (98.478) & 100.000 \\
\textbf{Test 10} & 99.389 & Unit Tests (100.00) Functional Tests (99.783) & 99.738 \\
\bottomrule
\end{tabular}}
\end{center}
\end{table*}

\begin{table*}
\begin{center}
\caption{The results of user study. The evaluation is divided into three levels: Satisfied/Needs Improvement/Unsatisfied, with user evaluation indicated in bold.}
\label{table:7}
\resizebox{0.65\linewidth}{!}{
\begin{tabular}{l|l|c}
\toprule
{\textbf{Num of User/Test}}& {\textbf{User evaluation}} & \textbf{Score} \\
\midrule
User 1 & \textbf{Satisfied}/Needs Improvement/Unsatisfied & 4.50 \\
User 2 & Satisfied/\textbf{Needs Improvement}/Unsatisfied & 4.25 \\
User 3 & \textbf{Satisfied}/Needs Improvement/Unsatisfied & 4.75 \\
User 4 & \textbf{Satisfied}/Needs Improvement/Unsatisfied & 5.00 \\
User 5 & \textbf{Satisfied}/Needs Improvement/Unsatisfied & 4.75 \\
\bottomrule
\end{tabular}}
\end{center}
\end{table*}

\begin{table*}
\begin{center}
\caption{Deployment options and provided computing resources.}
\label{table:8}
\resizebox{0.65\linewidth}{!}{
\begin{tabular}{l|l}
\toprule
{\textbf{Deployment Plan}}& {\textbf{Computing Resources}} \\
\midrule
\textbf{GitHub page+MkDocs} & Github hosting\\
\midrule
\multirow{2}{*}{\textbf{Server+Nginx}} & Plan 1: RTX3080  \\
        & Plan 2: 4 TITAN RTX\\
\midrule
\multirow{4}{*}{\textbf{Server+Parallel Computing}} & 32-core 2.5GHz X86 architecture processor\\
        & 4 GPU-like accelerator cards (16GB memory)\\
        & 128GB memory \\
        & mount shared storage system \\
\bottomrule
\end{tabular}}
\end{center}
\end{table*}

\begin{figure*}
    \centering
    \includegraphics[width=\textwidth]{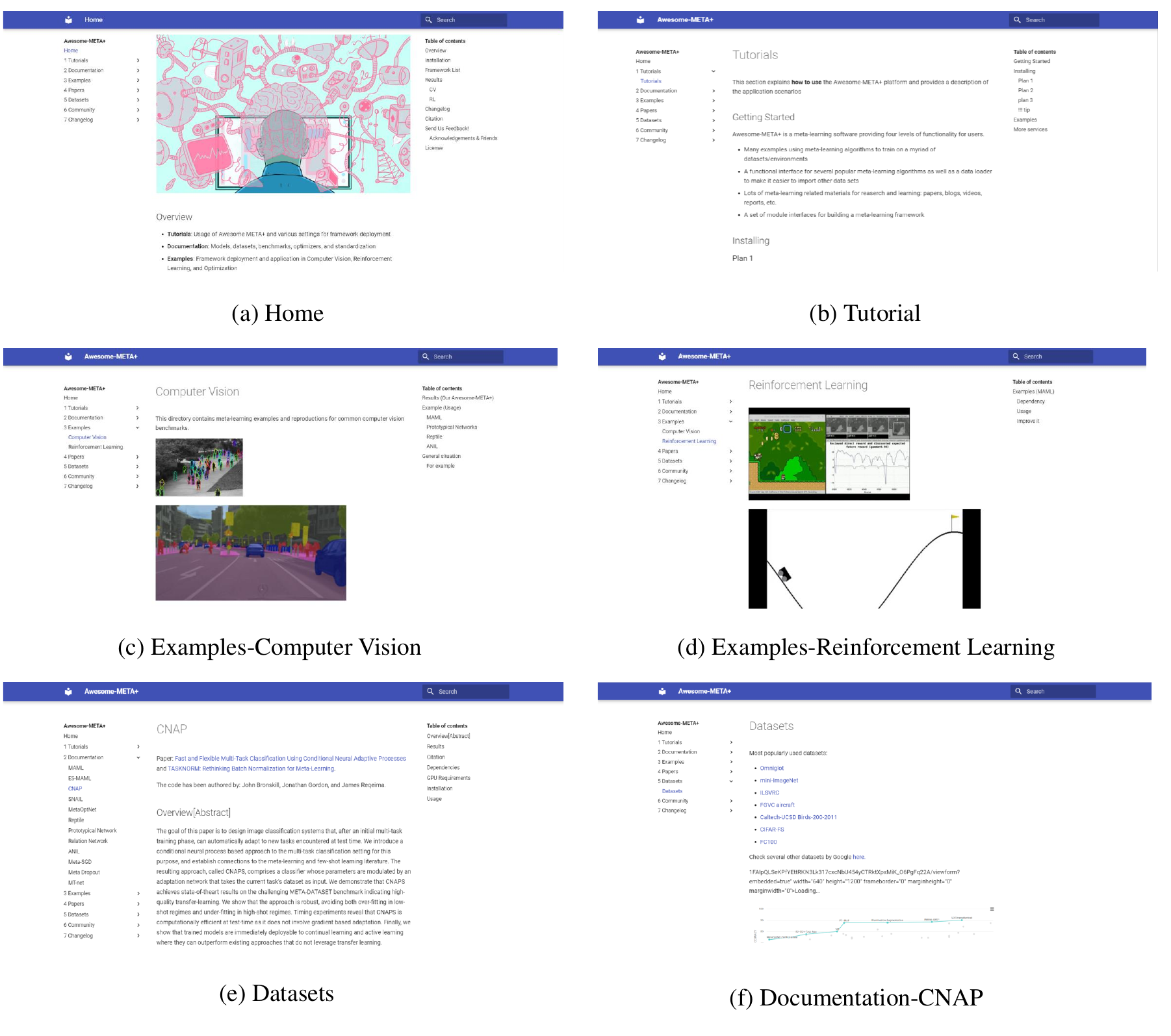}
    \caption{Examples of Awesome-META+.}
    \label{performance}
\end{figure*}


\section{Evaluation and Application}
\label{Verification}

\subsection{Testing and Verification}
To evaluate the effectiveness of the constructed AwesomeMeta+, we conduct machine-based testing and user studies. The former is used to assess the performance and responsiveness of AwesomeMeta+ when integrated into workflows and applications, while the latter evaluates whether AwesomeMeta+ could improve users' engineering efficiency, aid in understanding meta-learning, and guide future improvements. 

Specifically, the machine-based testing includes (i) web testing: function testing based on Playwright + Python for webpages, access and API response testing based on OctaGate SiteTimer, and (ii) model deployment testing: inter-module compatibility testing and response evaluation for given cases. The deployment is based on GitHub + Vercel to deploy the webpage demo (V1.0). Similarly, the user studies are also to evaluate whether AwesomeMeta+ can provide users with the required functions, but it is based on a questionnaire. The questionnaire includes the experience of using each module and the rating, which is provided in the Appendix.

\subsubsection{Machine-based Testing}
The machine-based testing involves web testing and model deployment testing.
Among them, web testing includes two components: testing the navigation and interaction of each interface and function, as well as performance testing for the response. The system provides nine interfaces, including "Home", "Tutorials", "Documentation", "Examples", "Papers", "Datasets", "Community", "Changelog", and "GitHub". Among them, "Tutorials", "Papers", "Datasets", and "Examples" can be linked to the "download" operation. "Tutorials" also involve online deployment, and the most important aspect is the navigation relationship between different interfaces. Therefore, we conduct testing from three core functional perspectives: "System Front-end Web Interaction", "System Framework Integration Testing and Deployment", and "Meta-Learning Information Acquisition". The results are shown in Table \ref{table:6}.

For system front-end web interaction, we chose to use the Python community's Playwright library for web testing. Playwright is a pure automation tool designed specifically for the Python language by Microsoft. It can automatically execute Chromium, Firefox, and WebKit browsers through a single API and can achieve automation functions while supporting headless and headed modes. Moreover, Playwright supports Linux, Mac, and Windows operating systems, making it a very suitable web testing tool for the Awesome-META+ system we want to build.

For integrated framework testing: we design each framework modularly, which can be activated or deactivated according to the needs. The middle of Table \ref{table:6} shows the corresponding results of each framework from unit testing to functional testing. The unit testing involves the whole project of each model and the standardized modules. The functional testing covers all algorithms provided by the system. Each case includes multiple datasets and tasks, and some cases contain multiple training modes (e.g., distributed and single-card training), and the core algorithms are encapsulated in the form of classes and functions.

For meta-learning information acquisition: we randomly extract 600 data items related to meta-learning, including paper keywords, title entity words, author, and conference information. 60\% of the test set is collected from the provided materials, and the remaining 40\% was selected from the paper search websites corresponding to the meta-learning-related entries. By inputting the test set data randomly and repeatably, setting different repeat extraction probabilities, and taking 300 times as a group, we calculate the average accuracy and response speed for a total of 10 groups to test the meta-learning information acquisition part. The results are shown in the right of Table \ref{table:6}.

\subsubsection{User Study}
To further ensure the effectiveness of system interaction in actual use, we conduct a user study. We provide the testing interface of AwesomeMeta+ to 50 volunteers and collect their feedback through a questionnaire after they have used it for one week. Notably, these 50 volunteers are the same as those in the formative study, as described in Section \ref{sec:3}. The user study consists of three phases. In each phase, we use a rating system ranging from 1 to 5, where scores of 4.5 and above are classified as "Satisfied," scores between 3.5 and 4.5 as "Needs Improvement," and scores below 3 as "Unsatisfied." Finally, we obtain the overall evaluation of AwesomeMeta+ by collecting the average score across the three phases.

Specifically, in the first phase, to avoid user experience issues caused by webpage errors (such as redirection failures), we manually test the front-end interaction and model deployment processes multiple times. During this process, users search for the required files and materials on their own, and we collect issues encountered during usage. Additionally, we set up error handling and guidance prompts for potential issues like empty form submissions, search failures, and ineffective model download controls, and record whether users can obtain useful information from these prompts. Second, we evaluate the model deployment. We provide five deployment cases, ask users to randomly select two of them for deployment, and then provide feedback on the deployment results, including performance and model effectiveness. Finally, in the third phase, we allow users to set their goals based on their engineering or research scenarios and provide feedback on whether AwesomeMeta+ can support or accelerate their progress. The usage results are collected from the users where part of them are presented in Table \ref{table:7}.

Multiple tests show that users can obtain resource download responses within 50ms without being affected by internet speed; empty form submissions and empty package downloads are not expected to occur; the search function can quickly locate the corresponding module. For model deployment operations, users are able to run the models smoothly on both the server and local machines for training following the deployment instructions.

\subsection{Deployment and Maintenance}
The web page of AwesomeMeta+ is deployed in two ways: GitHub Pages for the display system and server+nginx for the system usage. The specific deployment schemes and provided computing resources are presented in Table \ref{table:8}.

To ensure the long-term performance of Awesome-META+ and support the research functionalities, we design the system sustainably and reserve interfaces for future updates and iterations. The main work includes: (i) Modularization of functionalities: All frameworks and specific functionalities are designed with modularity, and ports for activation or deactivation are integrated into deployment instructions, making it easy to locate the functional blocks for future modifications and supplements. (ii) Developer-oriented comments are included in the code, and standardized documents and system design schemes are provided for other standardized engineering in different fields. (iii) The integrated frameworks, datasets, optimization packages, and academic materials are all packaged for upload, like building blocks that can be continuously supplemented on the basis of existing resources. With sufficient computing resources, there is no upper limit. (iv) The "Changelog" page of Awesome-META+ provides explanations for version iterations and updates.

\subsection{Performance Optimization}
The Awesome-META+ system consists of nine major modules, including "Home", "Tutorials", "Documentation", "Examples", "Papers", "Datasets", "Community", "Changelog", and "GitHub". These modules cover all the necessary aspects for applying typical meta-learning frameworks, such as deployment, usage tutorials, source code, and practical cases, as well as providing information on meta-learning academic resources, including papers, datasets, blogs, and video tutorials. Additionally, the system includes multiple modules related to system updates and community building. Figure \ref{performance} illustrates the effects of the Awesome-META+ system.


\section{Conclusion}
This study presents AwesomeMeta+, a prototyping and learning system designed to address the technical challenges associated with the deployment and integration of meta-learning within complex system environments. By standardizing, modularizing, and integrating diverse models, AwesomeMeta+ supports systematic model development, deployment, and optimization, facilitating more efficient decision-making in real-world applications. Through a formative study and iterative design refinements, we establish a structured framework that enhances the usability, adaptability, and scalability of meta-learning workflows. Machine-based testing and user studies demonstrate that AwesomeMeta+ accelerates system engineering processes, enabling users to effectively select, deploy, and manage meta-learning models with minimal overhead. Beyond meta-learning, this study serves as a generalizable example of model integration and standardization in AI-driven systems engineering, offering insights into improving model interoperability, deployment efficiency, and adaptive learning strategies across various domains.


\bibliographystyle{IEEEtran}
\bibliography{reference}


\end{document}